\pdfoutput=1

\documentclass[11pt]{article}
\usepackage[dvipsnames]{xcolor}

\usepackage[final]{acl}

\usepackage{times}
\usepackage{latexsym}

\usepackage[T1]{fontenc}

\usepackage[utf8]{inputenc}

\usepackage{microtype}

\usepackage{inconsolata}

\usepackage{graphicx}

\usepackage{xurl}

\usepackage{amsmath}
\usepackage{stmaryrd}
\usepackage{verbatim}

\usepackage{tabularx}
\usepackage{booktabs}
\newcolumntype{C}{>{\centering\arraybackslash}X}
\usepackage{multirow}
\usepackage{diagbox}
\usepackage{hhline}
\usepackage{color}
\usepackage{amsmath}
\usepackage{amssymb}
\usepackage{mathtools}
\usepackage{soul}

\definecolor{RoseQuartzBg}{HTML}{F7CAC9}
\definecolor{RoseQuartz}{HTML}{F5A798}
\definecolor{Serenity}{HTML}{92A8D1}
\definecolor{OrangeRed}{rgb}{1.0, 0.27, 0.0}
\definecolor{Turquoise}{HTML}{0F4C81}

\usepackage{arydshln}
\definecolor{themered}{HTML}{FF8375}

\usepackage{xparse}
\usepackage{bbm}
\usepackage{caption}
\usepackage{subcaption}
\usepackage{stfloats}
\usepackage{pifont}
\newcommand*\colourcheck[1]{
  \expandafter\newcommand\csname #1check\endcsname{\textcolor{#1}{\ding{52}}}
}
\newcommand*\colourxmark[1]{
  \expandafter\newcommand\csname #1xmark\endcsname{\textcolor{#1}{\ding{56}}}
}

\colourcheck{green}
\colourxmark{red}
\usepackage{cancel}
\usepackage{transparent}
\usepackage{float}
\usepackage{tcolorbox}
\tcbset{colframe=black,colback=white,size=small,colbacktitle=gray!30!white,coltitle=black,fonttitle=\bfseries,fontupper=\footnotesize\ttfamily}

\NewDocumentCommand{\haoyang}{ mO{} }{
\textcolor{Turquoise}{\textsuperscript{\textsc{Haoyang}}\textsf{\textbf{\small[#1]}}}}

\title{On Synthetic Data Strategies for Domain-Specific Generative Retrieval}

\author{Haoyang Wen$^\ddagger$, Jiang Guo$^\dagger$\thanks{Corresponding author}, Yi Zhang$^\dagger$, Jiarong Jiang$^\dagger$, Zhiguo Wang$^\dagger$\\
    $^\ddagger$Language Technologies Institute, Carnegie Mellon University~~~~$^\dagger$AWS AI\\
    \texttt{hwen3@cs.cmu.edu}\\
    \texttt{\{gujiang, imyi, jiarongj, zhiguow\}@amazon.com}}

\begin{document}
\maketitle
\begin{abstract}
This paper investigates synthetic data generation strategies in developing generative retrieval models for domain-specific corpora, thereby addressing the scalability challenges inherent in manually annotating in-domain queries. We study the data strategies for a two-stage training framework: in the first stage, which focuses on learning to decode document identifiers from queries, we investigate LLM-generated queries across multiple granularity (e.g. chunks, sentences) and domain-relevant search constraints that can better capture nuanced relevancy signals. In the second stage, which aims to refine document ranking through preference learning, we explore the strategies for mining hard negatives based on the initial model's predictions. Experiments on public datasets over diverse domains demonstrate the effectiveness of our synthetic data generation and hard negative sampling approach.
\end{abstract}
\section{Introduction}

Generative retrieval is emerging as a promising paradigm for information retrieval (IR), leveraging generative models (\textit{e.g.,} Transformers, \citealp{DBLP:conf/nips/VaswaniSPUJGKP17}) to directly produce ranked lists of potentially relevant document identifiers for a user query.
Although prior work has made progress on various fronts, including training strategies (\textit{e.g.,} identifier choices)~\citep{DBLP:conf/nips/Tay00NBM000GSCM22,DBLP:journals/corr/abs-2208-09257,DBLP:conf/nips/BevilacquaOLY0P22,DBLP:conf/nips/0001YCWZRCYRR23}, modeling techniques~\citep{chen-etal-2023-understanding,zhou-etal-2023-enhancing-generative,DBLP:conf/aaai/00010WWL24}, and inference methods~\citep{DBLP:journals/corr/abs-2010-00904,lee-etal-2023-nonparametric,DBLP:conf/sigir/ZhangL0DLC24}, the role of \emph{data strategies} in training generative retrieval models, particularly when dealing with domain-specific corpora, remains relatively underexplored.
This gap is critical: as generative retrieval models internalize entire corpus within their parametric memory, the choice and quality of training data are likely to play a critical role in their performance.

To mitigate the high cost and scalability challenges of in-domain annotation, most studies have adopted DSI-QG \cite{DBLP:journals/corr/abs-2206-10128}, which uses passage-level synthetic queries generated by docT5query \cite{nogueiradoc2query} (a model trained on MS-MARCO data).
However, applying such off-the-shelf synthetic data strategies to new domains may not suffice.
Unlike dense retrieval approaches, which focuses on strong text representation \cite{karpukhin2020dense,izacard2021contriever}, 
generative retriever must develop three key capabilities: (1) \textbf{memorization} (storing the content of the corpus (\textit{e.g.,} documents) and mapping them to their assigned identifiers), (2) \textbf{generalization} (inferring beyond explicit textual cues from user queries), and (3) \textbf{relevance scoring} (accurately ranking document identifiers by relevance to a given query).
Domain-specific corpora can amplify these challenges, as the model must adapt its internal representations to reflect domain nuances while maintaining robust generalization and ranking accuracy.
In this work, we systematically investigate data strategies that can foster these core capabilities.

We introduce a two-stage training framework.
The first stage focuses on mapping an input directly to document identifiers via supervised fine-tuning on synthetic data. The second stage uses preference learning to further enhance the ranking performance~\citep{zhou-etal-2023-enhancing-generative,DBLP:conf/aaai/00010WWL24}. Here we adopt Regularized Preference Optimization (\citealp[RPO]{DBLP:journals/corr/abs-2404-19733}), an effective alternative to PPO-based reinforcement learning \citep{DBLP:conf/nips/Ouyang0JAWMZASR22}.
We study the data strategies for both stages.

The first stage focuses on the memorization and generalization ability. 
We examine two data sources as the input for decoding document identifiers during training: the \emph{context} data (\textit{e.g.,} chunks) directly extracted from the corpus, and \emph{synthetic queries} that represent various relevance signals. For synthetic queries, we investigate query generation using multi-granular context (\textit{e.g.,} sentence-level, chunk-level) to capture both local and global information from the corpus. We also explore adding constraints derived from available metadata or domain-specific knowledge when generating synthetic queries to enhance the model's ability of handling complex, domain-relevant queries.

In the first stage, models are optimized to produce a single positive candidate, without considering relevance between different candidates. In the second stage, we further create data to enhance the model's ranking capability through preference learning~\citep{zhou-etal-2023-enhancing-generative,DBLP:conf/aaai/00010WWL24}. 
We study the selection of negative candidate documents for preference learning. Rather than relying on static offline data, we collect preference data online from the model's own top-ranked candidates after the first stage, and compare it to random sampling from the corpus.
We further investigate the choices and impact of varying the number of negative candidates to the ranking performance.

We conduct experiments on datasets covering various aspects of relevance, including the widely adopted Natural Questions (\citealp[NQ]{kwiatkowski-etal-2019-natural}), a multi-hop dataset MultiHop-RAG~\citep{DBLP:journals/corr/abs-2401-15391}, and two perspective-based retrieval datasets: AllSides~\citep{baly-etal-2020-detect} and AGNews~\citep{DBLP:conf/nips/YuZZMRKSZ23} from \citet{DBLP:journals/corr/abs-2405-02714}.
We show that queries with different aspects, such as multi-granular and constrains-based queries, significantly improve the retrieval performance compared to relying solely on chunk-level synthetic queries from query generation models. Additionally, upsampling context data further improves the performance.
Moreover, we show that these data strategies generalize well to other types of document identifiers, such as atomic identifiers.
Finally, we demonstrate that RPO effectively improve the ranking performance of generative retrieval, and the key lies in the selection of high-quality negative candidates: high-quality hard negative candidates improve the performance while random negatives may have an adverse impact.

In summary, this work offers a comprehensive investigation of data strategies for building scalable and effective domain-specific generative retrieval systems. Our findings emphasizes the importance of creating high-quality and diverse synthetic queries that capture multiple levels of granularity within the corpus, as well as informed negative selection strategies for ranking optimization.

\section{Generative Retrieval Framework}
A typical generative retrieval framework takes a query as input, and generates the corresponding relevant document identifiers as the retrieval results~\citep{DBLP:conf/nips/Tay00NBM000GSCM22}. Because each document in the corpus has a unique identifier, one can then use these identifiers to retrieve the corresponding documents for downstream tasks.

\subsection{Document Identifiers}
We primarily use \textit{semantic} document identifiers in our experiments due to their superior performance and better scalability to larger corpora.
Instead of using corpus-specific semantic identifiers like titles or urls, we adopt a more general, keyword-based approach that can be applied to a wide variety of corpora~\citep{DBLP:journals/corr/abs-2208-09257}.
Specifically, we instruct an LLM to produce a list of keywords that describes the content of a document, and use this keyword list as its semantic identifier.

In addition, we extended our synthetic data strategies to other types of identifiers to validate its generalizability, such as atomic identifiers ~\citep{DBLP:conf/nips/Tay00NBM000GSCM22}, which are unique tokens that can be generated through a one-step decoding or classification process.

\subsection{Generative Modeling}
The generative retrieval model learns to generate the identifier of a relevant document given a query. Formally, for a query $q$ and a relevant document $d$ with identifier $d'$, generative retrieval aims to produce $d'$ given $q$, which can be represented as:
\newcommand{\score}[1]{\operatorname{score}(#1)}
\begin{align*}
\score{q,d} &= P\left(d'\mid q; \theta\right) \\
&= \prod_{i}P\left(d'_i \mid d'_{<i}, q; \theta \right),
\end{align*}
where $d'_{i}$ is the $i^\text{th}$ token of the identifier. To ensure the generated identifiers are valid during inference, we use constrained beam search with Trie~\citep{DBLP:journals/corr/abs-2010-00904} to restrict the output token space at each decoding step. The top-$k$ output from the beam search serves as the final retrieval results.

Compared to dense retrieval models~\cite{karpukhin2020dense}, generative retrieval bypasses the need for an external index by directly producing relevant document identifiers. However, there are distinct challenges in learning a generative retrieval model.
As it solely relies on parametric knowledge, the model must not only learn the retrieval task, but also capture and encode document content in a way that associates each document with its identifier. Therefore, generative retrieval often requires training on the entire corpus to enable the model to memorize and comprehend the necessary information effectively. 

\begin{figure*}[htbp]
    \centering
    \includegraphics[width=\linewidth, trim={0.6cm 0.2cm 0.6cm 0.3cm}, clip]{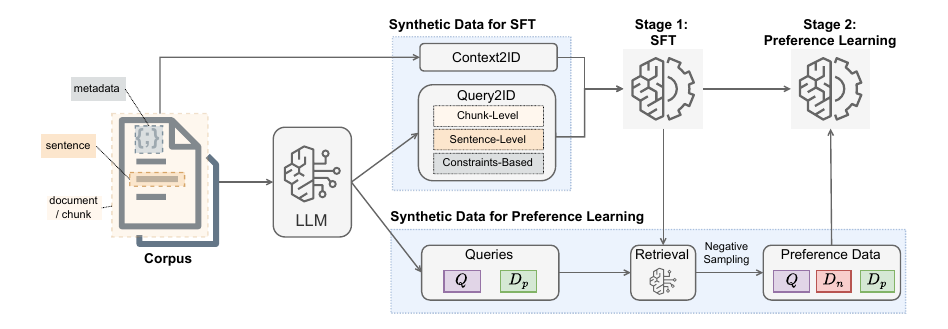}
    \caption{The overall workflow of the generative retrieval training and synthetic data utilization at each stage.}
    \label{fig:workflow}
\end{figure*}

\section{Supervised Fine-Tuning Data Strategy}
In a typical domain-specific setup, we often assume access to a corpus with limited or no labeled data for domain-specific training~\citep{DBLP:conf/ictir/HashemiZKPMC23}. Therefore, it is crucial to create high-quality synthetic data that thoroughly covers the entire corpus for generative retrieval training.

Our synthetic data comprises two main components: Context2ID data and Query2ID data. Context2ID involves training the model to retrieve the document identifiers given the document's content. Query2ID focuses on teaching the model to retrieve relevant document identifiers from a given query.
Combining these two data types encourages the model to learn both content memorization and retrieval given a query.

\subsection{Supervised Fine-Tuning Objective}
At this stage, we train the model to generate relevant document identifiers by maximizing the probability of each individual token. While typical supervised fine-tuning (SFT), especially with encoder-decoder architectures such as T5, focuses on optimizing the output sequence (\textit{i.e.} document identifiers), it's also part of the training goal for generative retrieval models to comprehend and memorize the context. To this end, we also optimize the model for learning to decode the input.
Specifically, for a given query-document pair $(q,d)$, where $q$ could be an actual query or a text chunk from the document, the model maximizes the likelihood of the combined input and output sequence:
\begin{align*}
\mathcal{L}_\text{sft}\left(q,d\right) = &-\log P\left(d', q; \theta\right) \\
= &-\sum_i \log P(q_i \mid q_{<i}; \theta) \\
 &- \sum_i \log P(d'_i \mid d'_{<i}, q; \theta).
\end{align*}

\begin{table*}[htbp]
    \centering
    \small
    \begin{tabular}{lp{0.749\textwidth}}
        \toprule
        \textbf{Data Type} & \textbf{Example} \\
        \midrule
        Context & title: Christmas Day preview: \colorbox{Apricot}{49ers}, \colorbox{Salmon}{Ravens} square off in potential Super Bowl sneak peek\ldots source: \colorbox{GreenYellow}{Yardbarker} \ldots \colorbox{Apricot}{San Francisco} has racked up an NFL-leading 25 turnovers and has given up the second-fewest rushing \colorbox{Goldenrod}{yards (1,252)}, \ldots \\
        Chunk-Level Query & What is the potential implication of this matchup between the \colorbox{Apricot}{49ers} and \colorbox{Salmon}{Ravens}? \\
        Sentence-Level Query &  Where does the \colorbox{Apricot}{49ers}' defense stand in terms of \colorbox{Goldenrod}{total yards} allowed per game? \\
        Constraints-Based Query & \underline{According to the \colorbox{GreenYellow}{Yardbarker} article}, which team has the league's most effective running game?\\
        \bottomrule
    \end{tabular}
    \caption{Examples of different synthetic queries generated from MultiHop-RAG corpus.} %
    \label{tab:data_type_example}
    \vspace{-0.5em}
\end{table*}

\subsection{Context2ID}
Context2ID data is created by pairing each chunk of text in the corpus with its corresponding document identifier. The goal of Context2ID data is to help the generative retrieval model associate each document's content with its unique identifier, i.e., ``memorizing'' the text.

\subsection{Query2ID}
Query2ID is designed to teach the model to retrieve the relevant document identifiers given a query. It helps the model to learn the core retrieval task and also further comprehend content from the query perspective.

Previous work~\citep{DBLP:journals/corr/abs-2206-10128} finds it effective to use a query generation model (\textit{e.g.,} docT5query, \citealp{nogueiradoc2query}) to produce synthetic queries from documents using multiple independent samplings. In this work, we instead use an LLM for synthetic query generation. Specifically, given a context (e.g., a document chunk), the LLM is instructed to generate a diverse set of $m$ queries, thereby covering a  wider range of semantic variations compared to the sampling-based approach with a specialized query generation model.

\subsubsection{Multi-Granular Query Generation} We first generate queries with context at different levels of granularity: \textit{chunk-level} and \textit{sentence-level}. Chunk-level synthetic queries are produced by providing the entire chunk as input to the LLM to capture higher-level semantics or facts, while sentence-level synthetic queries are produced by only providing individual sentences to focus on more specific details within the document
Concretely, for each chunk, we ask the LLM to produce $m_c$ chunk-level queries. We then split the chunk into individual sentences and ask the LLM to generate $m_s$ sentence-level queries for each sentence.

\subsubsection{Constraints-Based Query Generation}

A key advantage of using an LLM for query generation is its ability to incorporate domain-specific instructions.
For instance, we can prompt the LLM to include metadata constraints, such as the \textit{author name} or \textit{political polarity} of a document, in the generated queries.
Although the specific constraint types depends on the metadata available and can be domain or dataset specific, they are common in real-world scenarios such as enterprise data. Table~\ref{tab:dataset_attributes} in Appendix specifies the attributes that we use to produce constraints-based synthetic queries for each dataset. We ask the LLM to generate $m_i$ queries for each document that incorporate these constraints, allowing our generative retrieval model to handle more specialized or domain-specific queries.

\section{Preference Learning Data Strategy}
Previous work~\citep{zhou-etal-2023-enhancing-generative,DBLP:conf/aaai/00010WWL24} have shown that incorporating ranking tasks can further enhance the relevance modeling of generative retrieval models. However, when generative retrieval models are based on large language models, complex ranking objectives -- such as listwise optimization -- often become computationally inefficient due to multiple forward passes. In this work, we instead use a simpler method and adopt the regularized preference optimization algorithm to perform the preference optimization, a technique widely applied in optimizing large language models. We will first briefly introduce the preference optimization method, and then turn our focus on the synthetic data construction, which consists of the synthetic queries along with their corresponding preferred or rejected candidates.

\subsection{Preference Optimization Objective}
We use Regularized Preference Optimization (\citealp[RPO]{DBLP:journals/corr/abs-2404-19733}) as our optimization method for preference learning. It is an extended version of Directed Preference Optimization (\citealp[DPO]{DBLP:conf/nips/RafailovSMMEF23}), including additional supervised fine-tuning loss to alleviate the over-optimization issues on negative responses. It takes an input query $q$, a positive candidate $d_p$, and a negative candidate $d_n$ as input. The loss is in favor of the positive candidate while against the negative candidate
\begin{align*}
    \mathcal{L}_\text{rpo}\left(q, d_p, d_n\right) = &- \log \delta \left(  \beta\log \frac{P\left(d'_p\mid q; \theta\right)}{P\left(d'_p\mid q; \theta_\text{ref}\right)} \right.\\
    & \quad\quad\ \ \ \left. -\beta\log \frac{P\left(d'_n\mid q; \theta\right)}{P\left(d'_n\mid q; \theta_\text{ref}\right)} \right) \\
    & - \alpha\frac{\log P(d'_p \mid q; \theta)}{\left|d'_p\right|},
\end{align*}
where $\theta_\text{ref}$ is the parameter of the reference model, \textit{i.e.,} the supervised fine-tuned model from the first stage training. $d'_p$ and $d'_n$ are the identifiers of the positive and negative candidate, respectively.

\subsection{Synthetic Queries}
Similar to the previous section, in a domain-specific setup, we assume that we do not have enough data for model training. Therefore, after the supervised fine-tuning stage, we need a batch of new synthetic queries for preference learning.

We still adopt the LLM-based query generation as with the supervised fine-tuning stage. However, there are a few key differences in the instructions. First of all, we ask the LLM to make queries as difficult as possible. At the same time, we ask the LLM to provide not only the synthetic queries but also their corresponding answers. This is to ensure that, while making difficult queries, those synthetic queries are still answerable using the given context.

These changes make the new batch of synthetic queries different from queries used during supervised fine-tuning so that the model will not be over-optimized to the same batch of data. Intensifying the difficulties also increases the likelihood that the initial generative retrieval model makes mistakes, and therefore the model will benefit from the preference learning by learning from those mistakes.

\subsection{Candidate Selection}
After producing the synthetic queries, the next step is to select document candidate pairs for RPO optimization. For each training instance, we need one positive candidate and one negative candidate. As we always produce synthetic queries based on a document, the positive candidate can be naturally assigned. Therefore, the focus will be on selecting negative candidates for each synthetic query.

To increase the hardness of the negative candidates, we choose to select negative candidates from the retrieval results. Specifically, after the supervised fine-tuning stage, we will use the generative retrieval model to perform retrieval on the synthetic queries for preference learning. Our strategy mainly focuses on selecting the top-$k$ negative candidates with ranks higher than the positive candidate from the retrieval results. In this way, if the positive candidate ranks in the top-1, we will not use the query for preference learning. If the rank of the positive candidate is higher than $k$, then there will be different numbers of negative candidates, depending on the rank. If the rank is lower than $k$, there will be $k$ different negative candidates. When there are multiple negative candidates, we pair each negative candidate with the positive one to form a candidate pair instance for preference learning.

\section{Experiments}
\subsection{Datasets}
We choose 4 datasets for our experiments: three domain-specific corpora -- MultiHop-RAG~\citep{DBLP:journals/corr/abs-2401-15391}, --AllSides~\citep{baly-etal-2020-detect} and AGNews~\citep{DBLP:conf/nips/YuZZMRKSZ23} from \citet{DBLP:journals/corr/abs-2405-02714} -- as well as the general-domain dataset Natural Questions dataset~(\citealp[NQ]{kwiatkowski-etal-2019-natural}).

For AllSides and AGNews, we mainly adopt queries from \citet{DBLP:journals/corr/abs-2405-02714}.
In the case of AGNews, we replace the similar document part in queries with another attribute of perspective, as we focus on the query retrieval rather than document similarity search.

For NQ, we use the ``old document'' split from \citet{DBLP:conf/icml/KishoreWLAW23}, which constructs a subset of Wikipedia pages containing all positive candidates for training and testing, while keeping the corpus size manageable for generative retrieval training.

\subsection{Experiment Setup}
\label{sec:experiment_setup}
For all datasets, we use Mistral 7b~\citep{DBLP:journals/corr/abs-2310-06825} series as the generative retrieval base model. We use Mixtral 8x7b~\citep{DBLP:journals/corr/abs-2401-04088} to generate all the synthetic queries and we use Claude 3 Sonnet~\citep{anthropic2024claude} to generate keywords. Please refer to Appendix~\ref{app:data_specific_setup} for more details on the training infrastructure, hyperparameters, dataset-specific setup and statistics.

\begin{table}[t]
    \centering
    \small
    \resizebox{0.48\textwidth}{!}{
    \begin{tabular}{ccccc}
    \toprule
     & \textbf{HIT@4} & \textbf{HIT@10} & \textbf{MAP@10} & \textbf{MRR@10}\\
     \midrule
     Chunk &43.64& 66.65& 13.98& 31.14\\
     +Sent &61.64 &81.69 &22.13 &47.20 \\
     \bottomrule
    \end{tabular}
    }
    \caption{Ablation study on the effect of synthetic queries generated at a sentence-level granularity of context on Multihop-RAG.}
    \label{tab:granularity}
    \vspace{-0.8em}
\end{table}

\subsection{Results}
We will discuss our experiment results for each of the stages. In the supervised fine-tuning stage, we will discuss the effects of multi-granular synthetic queries, synthetic data with domain-specific constraints, and the use of Context2ID data. For the preference learning stage, we will discuss using different candidates for preference learning.

\subsubsection{Supervised Fine-Tuning Stage}

\paragraph{Effects of multi-granular synthetic queries.} We conduct an analysis on the effects of incorporating synthetic queries generated from the context at different levels of granularity on MultiHop-RAG. We train the generative retrieval model based on semantic identifiers on chunk-level Query2ID data (Chunk), comparing it with the model trained on chunk-level and sentence-level Query2ID data (+Sent), and both models use Context2ID data.
The results are shown in Table~\ref{tab:granularity}. We find that sentence-level synthetic queries can significantly improve retrieval performance, indicating that synthetic query generation with a small context can help capture more details from the document.

\begin{table*}[htbp]
    \centering
    \small
    \resizebox{\textwidth}{!}{
        \begin{tabular}{lccccccccccc}
        \toprule
        & \multicolumn{4}{c}{\textbf{MultiHop-RAG}} & \multicolumn{3}{c}{\textbf{AllSides}} &\multicolumn{3}{c}{\textbf{AGNews}}\\\cmidrule(lr){2-5}\cmidrule(lr){6-8}\cmidrule(lr){9-11}
        & \textbf{HIT@4} & \textbf{HIT@10} & \textbf{MAP@10} & \textbf{MRR@10} & \textbf{HIT@1} & \textbf{HIT@5} & \textbf{HIT@10} & \textbf{HIT@1} & \textbf{HIT@5} & \textbf{HIT@10}\\
        \midrule
        w/o constraints &61.64 &81.69 &22.13 &47.20 & 10.19 & 29.63 & 47.22 & 59.91& 83.94& 88.11\\
        w/ constraints &69.98& 88.34 &24.85 & 52.29 & 14.20& 38.58 & 51.85 & 62.19& 83.78& 88.24\\
        \bottomrule
        \end{tabular}
    }
    \caption{Ablation study on generative retrieval performances with or without the constraints-based synthetic queries.}
    \label{tab:constraints}
\end{table*}
\begin{table*}[htbp]

    \centering
    \small
    \begin{tabular}{lcccccccc}
    \toprule
    & \multicolumn{4}{c}{\textbf{MultiHop-RAG}} & \multicolumn{4}{c}{\textbf{Natural Questions}} \\\cmidrule(lr){2-5}\cmidrule(lr){6-9}
    & \textbf{HIT@4} & \textbf{HIT@10} & \textbf{MAP@10} & \textbf{MRR@10} & \textbf{HIT@1} & \textbf{HIT@5} & \textbf{HIT@10} & \textbf{MRR@10} \\
    \midrule
    w/o Context2ID &41.33 &69.31 &14.45 &31.25 & 69.72 & 85.58 & 89.01 & 76.57\\
    w/ Context2ID &69.98& 88.34 &24.85 & 52.29 & 70.71 & 86.48 & 89.85 & 77.54\\
    \bottomrule
    \end{tabular}
    \caption{Ablation study on generative retrieval performance trained with or without Context2ID data. The results demonstrate the helpfulness of Context2ID data and learning to memorize the context for generative retrieval.}
    \label{tab:context2id}
\end{table*}

\paragraph{Effects of constraints-based synthetic queries.} We further study the use of constraints-based synthetic queries that are customized for each domain-specific setting. We conduct experiments on three domain-specific corpora, MultiHop-RAG, AllSides, and AGNews. We compare the semantic identifier-based generative retrieval model trained with or without constraints-based synthetic queries, combing with the corresponding Context2ID data. The results are shown in Table~\ref{tab:constraints}. The results show that constraints-based synthetic queries can further improve retrieval performance, indicating that it is helpful to use LLM-produced synthetic queries for domain customization.

\begin{table}[t]
    \centering
    \small
    \resizebox{.48\textwidth}{!}{
        \begin{tabular}{lcccc}
        \toprule
         & \textbf{HIT@4} & \textbf{HIT@10} & \textbf{MAP@10} & \textbf{MRR@10}\\
         \midrule
         Concat &44.30& 72.77& 15.64& 33.59\\
         Interleave &69.98& 88.34 &24.85 & 52.29 \\
         \bottomrule
        \end{tabular}
    }
    \caption{Analysis on different ways of combining Query2ID and Context2ID data on Multihop-RAG. We compare simple concatenation (Concat) and interleaving (Interleave) that inherently upsamples the Context2ID data.} %
    \label{tab:context_concat}
    \vspace{-0.5em}
\end{table}

\paragraph{Effects of Context2ID data.} Existing work~\citep{DBLP:journals/corr/abs-2206-10128,DBLP:conf/icml/KishoreWLAW23} debates whether Context2ID data are useful for generative retrieval training. In this work, we consider Context2ID data as an important part of the data recipe, and also include the memorization of the context as part of the supervised fine-tuning objective. Therefore, we conduct an analysis that removes the Context2ID data on MultiHop-RAG and NQ, and the results are shown in Table~\ref{tab:context2id}. We can find that Context2ID data consistently improve generative retrieval performance. We also include the comparison of the strategies to combine Query2ID and Context2ID data, including simple concatenation or interleaving Context2ID and Query2ID that will upsample Context2ID data (\textit{i.e.,} smaller size dataset) on MultiHop-RAG in Table~\ref{tab:context_concat}, again illustrating the importance of Context2ID and learning to memorize context may strengthen the effects on Context2ID.

\begin{table*}[htbp]
    \centering
    \small
    \begin{tabular}{lcccccccc}
    \toprule
    & \multicolumn{4}{c}{\textbf{MultiHop-RAG}} & \multicolumn{4}{c}{\textbf{Natural Questions}} \\\cmidrule(lr){2-5}\cmidrule(lr){6-9}
     & \textbf{HIT@4} & \textbf{HIT@10} & \textbf{MAP@10} & \textbf{MRR@10}& \textbf{HIT@1} & \textbf{HIT@5} & \textbf{HIT@10} & \textbf{MRR@10} \\
     \midrule
     docT5query &50.86& 73.30& 17.60& 37.73 & 63.30 & 79.12 & 85.18 & 70.30\\
     Mixtral 8x7b &61.64 &81.69 &22.13 &47.20 & 70.71 & 86.48 & 89.85 & 77.54\\
     \bottomrule
    \end{tabular}
    \caption{Generative retrieval performance with synthetic queries from Mixtral 8x7b and docT5query. The results show that queries from Mixtral 8x7b can help train a better generative retrieval model.}
    \label{tab:qg_model}
\end{table*}

\begin{figure}[t]
    \centering
    \includegraphics[width=\linewidth]{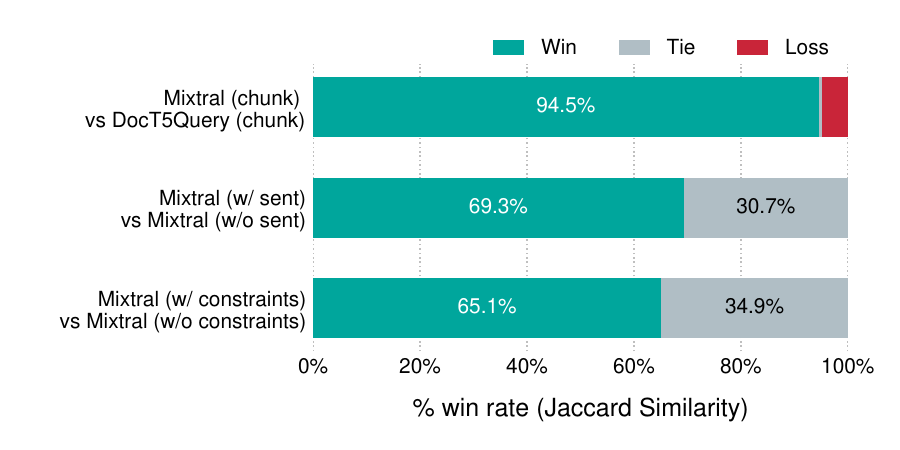}
    \caption{Jaccard similarity post-analysis on MultiHop-RAG test set. Synthetic queries from Mixtral 8x7b are generally closer to the test set than those from docT5query. Besides, incorporating granularity and domain-specific attributes further helps with getting queries that are closer to the test set.}
    \label{fig:jaccard_similarity}
    \vspace{-0.5em}
\end{figure}

\paragraph{Different query generation models.} As we primarily use LLMs to produce synthetic queries, it is important to understand the performance and effects of using an LLM compared to a specialized query generation model. To this end, we conduct a comparison between synthetic queries generated by Mixtral 8x7b and those from docT5query, as shown in Table~\ref{tab:qg_model}. For a fair comparison, we exclude constraints-based queries because docT5query cannot produce these types of queries. The results show generative retrieval models trained with queries from Mixtral 8x7b consistently outperform models trained on queries from docT5query.

Following \citet{pradeep-etal-2023-generative}, we use Jaccard similarity to evaluate the semantic similarity between test queries and synthetic queries as a post-analysis. 
For each test query, we compute the maximum Jaccard similarity among all synthetic queries generated for the corresponding chunk or document. We then compute the win rate between two synthetic query sets (e.g., Mixtral 8x7b versus docT5query) as the proportion of test queries for which one set exhibits higher Jaccard similarity.
Let \(Q\) be the set of test queries. Then, the win rate for a synthetic query set \(\mathcal{S}\) over another set \(\mathcal{T}\) is defined as:
\begin{align*}
& \text{Win Rate}(\mathcal{S}, \mathcal{T}) = \\
& \frac{1}{|Q|} \sum_{q \in Q} \mathbb{I}\left[\max_{s \in \mathcal{S}(q)} J(q, s) > \max_{t \in \mathcal{T}(q)} J(q, t)\right],
\end{align*}
where $J(q, s)$ denotes the Jaccard similarity between the token sequences of the test query $q$ and a synthetic query $s$, and $\mathcal{S}(q)$ represents the subset of synthetic queries for the corresponding chunk.

Figure~\ref{fig:jaccard_similarity} illustrates that synthetic queries from Mixtral 8x7b generally have a higher similarity to test queries. Moreover, both sentence-level synthetic queries and constraints-based queries contribute to improved distribution matching with the test queries.

\begin{table}[t]
    \centering
    \small
    \resizebox{.48\textwidth}{!}{
        \begin{tabular}{lcccc}
        \toprule
         & \textbf{HIT@4} & \textbf{HIT@10} & \textbf{MAP@10} & \textbf{MRR@10}\\
         \midrule
         all &74.32& 88.03& 29.71& 59.26\\
         \hdashline
         w/o Context2ID& 72.15& 86.21& 28.54&57.50\\
         w/o Sent &58.40&75.17&21.51&44.76\\
         w/o constraints &68.34&83.73&26.28&53.83\\
         \bottomrule
        \end{tabular}
    }
    \caption{Ablation study on atomic identifier-based generative retrieval performance on MultiHop-RAG.}
    \label{tab:identifier}
\end{table}
\paragraph{Generalization to different identifiers.} We further study the generalizability of our data strategies across different types of document identifiers. In this analysis, we use atomic identifier, which are arbitrary unique IDs assigned to each document or chunk. We conduct experiments on MultiHop-RAG and the results are shown in Table~\ref{tab:identifier}. The findings align with our observations using semantic identifiers, highlighting the critical role of all three data types in generative retrieval. Among them, sentence-level synthetic queries contribute the most to performance improvements.

\begin{table*}[htbp]
    \centering
    \small
    \begin{tabular}{lcccccccc}
    \toprule
    & \multicolumn{4}{c}{\textbf{MultiHop-RAG}} & \multicolumn{4}{c}{\textbf{Natural Questions}} \\\cmidrule(lr){2-5}\cmidrule(lr){6-9}
     & \textbf{HIT@4} & \textbf{HIT@10} & \textbf{MAP@10} & \textbf{MRR@10}& \textbf{HIT@1} & \textbf{HIT@5} & \textbf{HIT@10} & \textbf{MRR@10} \\
     \midrule
     SFT & 69.98& 88.34 &24.85 & 52.29 & 70.71 & 86.48 & 89.85 & 77.54 \\
     Random 5 & 58.94 & 82.88 & 20.88 & 43.53 & 70.19 & 86.48 & 89.50 & 77.17 \\
     Top-5 negative & 71.53 & 89.62 & 26.36 & 55.40 & 71.02 & 87.32 & 90.04 & 78.02 \\
     Top-10 negative & 71.88 & 89.80 & 26.23 & 54.94 & 71.22 & 87.41 & 89.97 & 78.14 \\
     \bottomrule
    \end{tabular}
    \caption{Preference learning with different numbers of negative candidates. The results show that it is an effective strategy to select negative candidates with ranks higher than the positive candidate, while different numbers of negative candidates may optimize the retrieval performance in different ways.}
    \label{tab:neg_cand_number}
\end{table*}

\subsubsection{Preference Learning Stage}

\paragraph{Effects of negative candidate sources.} We first study the strategies of candidates selection for preference learning. We compare random selection from the corpus with using the top candidates from the generative retrieval model after the supervised fine-tuning stage. The results are shown in Table~\ref{tab:neg_cand_number}, which illustrates that candidate selection has an impact on preference learning, and random candidates may have a negative impact.

\paragraph{Effects of negative candidate number.} We also study the effects of using different negative candidate numbers for each query. We experiment with selecting Top-5 and Top-10 negative candidates with a rank higher than the positive candidate from the retrieval results. The results are shown in Table~\ref{tab:neg_cand_number}. In general, it is effective to use the strategy, which includes high-quality candidates with ranks higher than the corresponding positive candidates. We also see some slight differences when including different numbers of negative candidates. We can find that a large number of negative candidates helps better in metrics such as HIT@1 and HIT@4.

\subsubsection{Comparison to Off-The-Shelf Retrievers}
\begin{figure*}[htbp]
    \centering
    \includegraphics[width=.9\linewidth, trim={0cm 0.5cm 0cm 0.3cm}, clip]{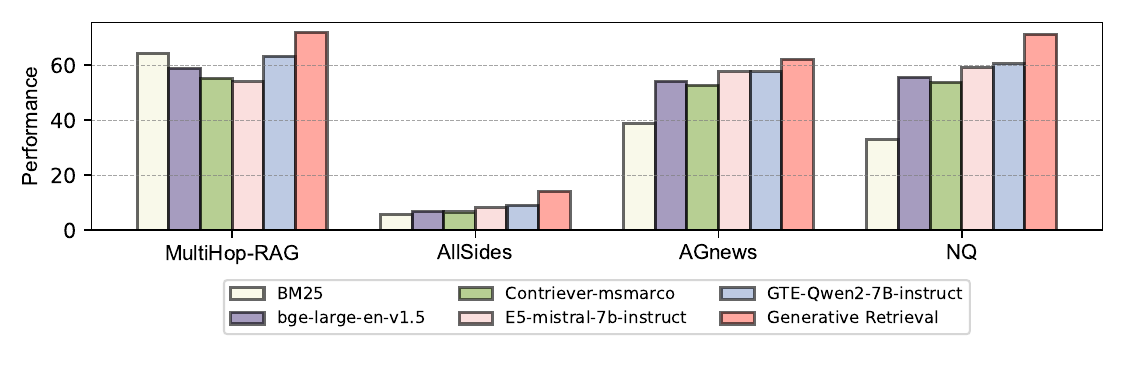}
    \caption{Performance comparison between generative retrieval using semantic identifiers and off-the-shelf-retrieval models. We use HIT@4 for MultiHop-RAG and HIT@1 for other datasets as the evaluation metric. For the full experiment results, please refer to Appendix \ref{sec:detailed_dense_comparison} (Table \ref{tab:dense_retrieval_comparison}).}
    \label{fig:dense_comparison}
\end{figure*}
We also compare the performance of our generative retrieval with several off-the-shelf retrievers, including BM25~\citep{DBLP:conf/sigir/RobertsonW94}, bge-large-en-v1.5~\citep{bge_embedding},  Contriever-msmarco~\citep{izacard2021contriever},  E5-mistral-7b-instruct~\citep{wang-etal-2024-improving-text} and GTE-Qwen2-7B-instruct~\citep{li2023towards}. The results are shown in Figure~\ref{fig:dense_comparison} and more detailed results can be found in Appendix~\ref{sec:detailed_dense_comparison}. The results demonstrate that generative retrieval models, which rely solely on in-domain synthetic data training without retrieval pre-training, can achieve competitive performance compared to those retrievers.
These results indicate the potential of generative retrieval, as well as the effectiveness of using LLMs to generate synthetic data tailored to domain-specific needs.

\section{Related Work}

\paragraph{Generative retrieval modeling.}
Previous work has explored various aspects of generative retrieval. One line of research aims to find appropriate document identifiers for generation, such as numerical or atomic identifier~\citep{DBLP:conf/nips/Tay00NBM000GSCM22,DBLP:journals/corr/abs-2206-10128,DBLP:journals/corr/abs-2208-09257}, N-grams~\citep{DBLP:conf/nips/BevilacquaOLY0P22,DBLP:conf/sigir/Chen0GR0FC23}, titles or URLs~\citep{DBLP:journals/ipm/LiYWWL23,lee-etal-2022-generative,DBLP:conf/cikm/ChenZG0FC22,ziems-etal-2023-large}, keywords-based or summary-based semantic identifiers~\citep{lee-etal-2023-glen,DBLP:conf/kdd/Tang0GCZWYC23}, codebook~\citep{DBLP:conf/nips/ZhangWCCZMHDMWP23,yang-etal-2023-auto,DBLP:conf/www/Zeng0JSWZ24}, and full passages themselves~\citep{tang2024selfretrievalendtoendinformationretrieval}. There are also efforts to combine the advantages of different identifiers~\citep{li-etal-2023-multiview}. Another line of work tackles the optimization of generative retrieval models, such as incorporating ranking losses~\citep{zhou-etal-2023-enhancing-generative,DBLP:conf/aaai/00010WWL24,DBLP:journals/tois/TangZGRCC24}, or using auxiliary tasks to enhance training~\citep{DBLP:conf/sigir/LiD0L24}. During retrieval, different constrained decoding methods have been explored to obtain valid identifiers, such as FM-Index~\citep{DBLP:conf/nips/BevilacquaOLY0P22}, Trie-based~\citep{DBLP:journals/corr/abs-2010-00904}, and set-based inference~\citep{DBLP:journals/tois/TangZGRCC24}.

\paragraph{Synthetic query generation.}
Alongside the progress in generative retrieval modeling and optimization, synthetic query generation has emerged as a pivotal technique for enhancing retrieval systems, particularly in domains with limited annotated data.
In dense retrieval, synthetic queries have been used extensively to improve cross-domain performance.
For instance, \citet{ma2021zero} generated synthetic questions for target-domain documents with a question generation model trained on general-domain data, thereby improving retrieval performance in zero-shot settings.
Similarly, \citet{wang2022gpl} introduced generative pseudo labeling, which combines query generation with pseudo labeling using a cross-encoder to capture finer-grained ranking signals.
Further advancements include \citet{bonifacio2022inpars} and \citet{jeronymo2023inpars}, which leverage large language models to generate synthetic queries in a few-shot manner, and then combine with top K documents ranked by the conditional question generation probability, to train a domain-specific reranker.

Despite these successes in dense retrieval, the potential of synthetic data for generative retrieval has been underexplored.
Existing studies typically rely on passage-level synthetic queries generated by docT5query \cite{nogueiradoc2query}, following the DSI-QG paradigm \cite{DBLP:journals/corr/abs-2206-10128}.
\citet{chen-etal-2023-understanding} explores breaking documents into text fragments for query generation and memorization. However, there still lacks a comprehensive discussions on effective strategies for generating synthetic data tailored to domain-specific corpora, especially with LLMs.
This work investigates data strategies from multiple perspectives, including the generation of synthetic queries using multi-granularity contexts, incorporating search constraints, and exploring the impact of context data. For preference learning, \citet{zhou-etal-2023-enhancing-generative} proposes using preference learning objectives for generative retrieval with specialized reward models, though acquiring such models in a domain-specific setting can be challenging.
In contrast, our proposed preference learning strategy directly uses the retrieval results to obtain the preference data, offering a more streamlined approach for domain-specific applications.

\section{Conclusion}
In this work, we explore several strategies to produce synthetic data for generative retrieval training. We find that adding queries in multi-granularity and queries with domain-specific constraints can largely improve the generative retrieval performance during supervised fine-tuning, and memorizing document contents can also contribute to the generative retrieval training. We also find that it is critical to choose high-quality hard negative candidates to effectively use the preference learning objectives to further improve generative retrieval.

\section*{Limitations}
Our proposed synthetic data strategies mainly focus on the supervised fine-tuning and preference learning stage. But there are also settings that can largely improve the usability of generative retrieval, such as incremental learning or generalization to unseen documents. It is also important to extend the data strategy exploration for these settings. In addition, similar data strategies may also be effectively used to enhance dense retrieval domain adaptation. Further systematic research is needed to investigate the strategies for dense retrieval model fine-tuning, as well as the differences between generative and dense model training.

Our synthetic queries are mainly based on one document. However, queries from the real world may be more complex, such as those involving multiple documents with multi-hop reasoning or multi-evidence comparison. It is still under-investigation that to generate those complex queries and to use those queries during retrieval model training.

\bibliography{anthology,custom}

\begin{thebibliography}{57}
\providecommand{\natexlab}[1]{#1}

\bibitem[{Anthropic(2024)}]{anthropic2024claude}
AI~Anthropic. 2024.
\newblock The claude 3 model family: Opus, sonnet, haiku.
\newblock \emph{Claude-3 Model Card}, 1.

\bibitem[{Baly et~al.(2020)Baly, Da~San~Martino, Glass, and Nakov}]{baly-etal-2020-detect}
Ramy Baly, Giovanni Da~San~Martino, James Glass, and Preslav Nakov. 2020.
\newblock \href {https://doi.org/10.18653/v1/2020.emnlp-main.404} {We can detect your bias: Predicting the political ideology of news articles}.
\newblock In \emph{Proceedings of the 2020 Conference on Empirical Methods in Natural Language Processing (EMNLP)}, pages 4982--4991, Online. Association for Computational Linguistics.

\bibitem[{Bevilacqua et~al.(2022)Bevilacqua, Ottaviano, Lewis, Yih, Riedel, and Petroni}]{DBLP:conf/nips/BevilacquaOLY0P22}
Michele Bevilacqua, Giuseppe Ottaviano, Patrick S.~H. Lewis, Scott Yih, Sebastian Riedel, and Fabio Petroni. 2022.
\newblock \href {http://papers.nips.cc/paper\_files/paper/2022/hash/cd88d62a2063fdaf7ce6f9068fb15dcd-Abstract-Conference.html} {Autoregressive search engines: Generating substrings as document identifiers}.
\newblock In \emph{Advances in Neural Information Processing Systems 35: Annual Conference on Neural Information Processing Systems 2022, NeurIPS 2022, New Orleans, LA, USA, November 28 - December 9, 2022}.

\bibitem[{Bonifacio et~al.(2022)Bonifacio, Abonizio, Fadaee, and Nogueira}]{bonifacio2022inpars}
Luiz Bonifacio, Hugo Abonizio, Marzieh Fadaee, and Rodrigo Nogueira. 2022.
\newblock Inpars: Data augmentation for information retrieval using large language models.
\newblock \emph{arXiv preprint arXiv:2202.05144}.

\bibitem[{Cao et~al.(2020)Cao, Izacard, Riedel, and Petroni}]{DBLP:journals/corr/abs-2010-00904}
Nicola~De Cao, Gautier Izacard, Sebastian Riedel, and Fabio Petroni. 2020.
\newblock \href {https://arxiv.org/abs/2010.00904} {Autoregressive entity retrieval}.
\newblock \emph{CoRR}, abs/2010.00904.

\bibitem[{Chen et~al.(2023{\natexlab{a}})Chen, Zhang, Guo, de~Rijke, Liu, Fan, and Cheng}]{DBLP:conf/sigir/Chen0GR0FC23}
Jiangui Chen, Ruqing Zhang, Jiafeng Guo, Maarten de~Rijke, Yiqun Liu, Yixing Fan, and Xueqi Cheng. 2023{\natexlab{a}}.
\newblock \href {https://doi.org/10.1145/3539618.3591631} {A unified generative retriever for knowledge-intensive language tasks via prompt learning}.
\newblock In \emph{Proceedings of the 46th International {ACM} {SIGIR} Conference on Research and Development in Information Retrieval, {SIGIR} 2023, Taipei, Taiwan, July 23-27, 2023}, pages 1448--1457. {ACM}.

\bibitem[{Chen et~al.(2022)Chen, Zhang, Guo, Liu, Fan, and Cheng}]{DBLP:conf/cikm/ChenZG0FC22}
Jiangui Chen, Ruqing Zhang, Jiafeng Guo, Yiqun Liu, Yixing Fan, and Xueqi Cheng. 2022.
\newblock \href {https://doi.org/10.1145/3511808.3557271} {Corpusbrain: Pre-train a generative retrieval model for knowledge-intensive language tasks}.
\newblock In \emph{Proceedings of the 31st {ACM} International Conference on Information {\&} Knowledge Management, Atlanta, GA, USA, October 17-21, 2022}, pages 191--200. {ACM}.

\bibitem[{Chen et~al.(2023{\natexlab{b}})Chen, Liu, He, Sun, and Sun}]{chen-etal-2023-understanding}
Xiaoyang Chen, Yanjiang Liu, Ben He, Le~Sun, and Yingfei Sun. 2023{\natexlab{b}}.
\newblock \href {https://doi.org/10.18653/v1/2023.findings-acl.681} {Understanding differential search index for text retrieval}.
\newblock In \emph{Findings of the Association for Computational Linguistics: ACL 2023}, pages 10701--10717, Toronto, Canada. Association for Computational Linguistics.

\bibitem[{Dao(2024)}]{DBLP:conf/iclr/Dao24}
Tri Dao. 2024.
\newblock \href {https://openreview.net/forum?id=mZn2Xyh9Ec} {Flashattention-2: Faster attention with better parallelism and work partitioning}.
\newblock In \emph{The Twelfth International Conference on Learning Representations, {ICLR} 2024, Vienna, Austria, May 7-11, 2024}. OpenReview.net.

\bibitem[{Gugger et~al.(2022)Gugger, Debut, Wolf, Schmid, Mueller, Mangrulkar, Sun, and Bossan}]{accelerate}
Sylvain Gugger, Lysandre Debut, Thomas Wolf, Philipp Schmid, Zachary Mueller, Sourab Mangrulkar, Marc Sun, and Benjamin Bossan. 2022.
\newblock Accelerate: Training and inference at scale made simple, efficient and adaptable.
\newblock \url{https://github.com/huggingface/accelerate}.

\bibitem[{Hashemi et~al.(2023)Hashemi, Zhuang, Kothur, Prasad, Meij, and Croft}]{DBLP:conf/ictir/HashemiZKPMC23}
Helia Hashemi, Yong Zhuang, Sachith Sri~Ram Kothur, Srivas Prasad, Edgar Meij, and W.~Bruce Croft. 2023.
\newblock \href {https://doi.org/10.1145/3578337.3605127} {Dense retrieval adaptation using target domain description}.
\newblock In \emph{Proceedings of the 2023 {ACM} {SIGIR} International Conference on Theory of Information Retrieval, {ICTIR} 2023, Taipei, Taiwan, 23 July 2023}, pages 95--104. {ACM}.

\bibitem[{Izacard et~al.(2021)Izacard, Caron, Hosseini, Riedel, Bojanowski, Joulin, and Grave}]{izacard2021contriever}
Gautier Izacard, Mathilde Caron, Lucas Hosseini, Sebastian Riedel, Piotr Bojanowski, Armand Joulin, and Edouard Grave. 2021.
\newblock \href {https://doi.org/10.48550/ARXIV.2112.09118} {Unsupervised dense information retrieval with contrastive learning}.

\bibitem[{Jeronymo et~al.(2023)Jeronymo, Bonifacio, Abonizio, Fadaee, Lotufo, Zavrel, and Nogueira}]{jeronymo2023inpars}
Vitor Jeronymo, Luiz Bonifacio, Hugo Abonizio, Marzieh Fadaee, Roberto Lotufo, Jakub Zavrel, and Rodrigo Nogueira. 2023.
\newblock Inpars-v2: Large language models as efficient dataset generators for information retrieval.
\newblock \emph{arXiv preprint arXiv:2301.01820}.

\bibitem[{Jiang et~al.(2023)Jiang, Sablayrolles, Mensch, Bamford, Chaplot, de~Las~Casas, Bressand, Lengyel, Lample, Saulnier, Lavaud, Lachaux, Stock, Scao, Lavril, Wang, Lacroix, and Sayed}]{DBLP:journals/corr/abs-2310-06825}
Albert~Q. Jiang, Alexandre Sablayrolles, Arthur Mensch, Chris Bamford, Devendra~Singh Chaplot, Diego de~Las~Casas, Florian Bressand, Gianna Lengyel, Guillaume Lample, Lucile Saulnier, L{\'{e}}lio~Renard Lavaud, Marie{-}Anne Lachaux, Pierre Stock, Teven~Le Scao, Thibaut Lavril, Thomas Wang, Timoth{\'{e}}e Lacroix, and William~El Sayed. 2023.
\newblock \href {https://doi.org/10.48550/ARXIV.2310.06825} {Mistral 7b}.
\newblock \emph{CoRR}, abs/2310.06825.

\bibitem[{Jiang et~al.(2024)Jiang, Sablayrolles, Roux, Mensch, Savary, Bamford, Chaplot, de~Las~Casas, Hanna, Bressand, Lengyel, Bour, Lample, Lavaud, Saulnier, Lachaux, Stock, Subramanian, Yang, Antoniak, Scao, Gervet, Lavril, Wang, Lacroix, and Sayed}]{DBLP:journals/corr/abs-2401-04088}
Albert~Q. Jiang, Alexandre Sablayrolles, Antoine Roux, Arthur Mensch, Blanche Savary, Chris Bamford, Devendra~Singh Chaplot, Diego de~Las~Casas, Emma~Bou Hanna, Florian Bressand, Gianna Lengyel, Guillaume Bour, Guillaume Lample, L{\'{e}}lio~Renard Lavaud, Lucile Saulnier, Marie{-}Anne Lachaux, Pierre Stock, Sandeep Subramanian, Sophia Yang, Szymon Antoniak, Teven~Le Scao, Th{\'{e}}ophile Gervet, Thibaut Lavril, Thomas Wang, Timoth{\'{e}}e Lacroix, and William~El Sayed. 2024.
\newblock \href {https://doi.org/10.48550/ARXIV.2401.04088} {Mixtral of experts}.
\newblock \emph{CoRR}, abs/2401.04088.

\bibitem[{Karpukhin et~al.(2020)Karpukhin, O{\u{g}}uz, Min, Lewis, Wu, Edunov, Chen, and Yih}]{karpukhin2020dense}
Vladimir Karpukhin, Barlas O{\u{g}}uz, Sewon Min, Patrick Lewis, Ledell Wu, Sergey Edunov, Danqi Chen, and Wen~Tau Yih. 2020.
\newblock Dense passage retrieval for open-domain question answering.
\newblock In \emph{2020 Conference on Empirical Methods in Natural Language Processing, EMNLP 2020}, pages 6769--6781. Association for Computational Linguistics (ACL).

\bibitem[{Kishore et~al.(2023)Kishore, Wan, Lovelace, Artzi, and Weinberger}]{DBLP:conf/icml/KishoreWLAW23}
Varsha Kishore, Chao Wan, Justin Lovelace, Yoav Artzi, and Kilian~Q. Weinberger. 2023.
\newblock \href {https://proceedings.mlr.press/v202/kishore23a.html} {Incdsi: Incrementally updatable document retrieval}.
\newblock In \emph{International Conference on Machine Learning, {ICML} 2023, 23-29 July 2023, Honolulu, Hawaii, {USA}}, volume 202 of \emph{Proceedings of Machine Learning Research}, pages 17122--17134. {PMLR}.

\bibitem[{Kwiatkowski et~al.(2019)Kwiatkowski, Palomaki, Redfield, Collins, Parikh, Alberti, Epstein, Polosukhin, Devlin, Lee, Toutanova, Jones, Kelcey, Chang, Dai, Uszkoreit, Le, and Petrov}]{kwiatkowski-etal-2019-natural}
Tom Kwiatkowski, Jennimaria Palomaki, Olivia Redfield, Michael Collins, Ankur Parikh, Chris Alberti, Danielle Epstein, Illia Polosukhin, Jacob Devlin, Kenton Lee, Kristina Toutanova, Llion Jones, Matthew Kelcey, Ming-Wei Chang, Andrew~M. Dai, Jakob Uszkoreit, Quoc Le, and Slav Petrov. 2019.
\newblock \href {https://doi.org/10.1162/tacl_a_00276} {Natural questions: A benchmark for question answering research}.
\newblock \emph{Transactions of the Association for Computational Linguistics}, 7:452--466.

\bibitem[{Lee et~al.(2023{\natexlab{a}})Lee, Kim, Chang, Oh, Yang, Karpukhin, Lu, and Seo}]{lee-etal-2023-nonparametric}
Hyunji Lee, JaeYoung Kim, Hoyeon Chang, Hanseok Oh, Sohee Yang, Vladimir Karpukhin, Yi~Lu, and Minjoon Seo. 2023{\natexlab{a}}.
\newblock \href {https://doi.org/10.18653/v1/2023.findings-acl.801} {Nonparametric decoding for generative retrieval}.
\newblock In \emph{Findings of the Association for Computational Linguistics: ACL 2023}, pages 12642--12661, Toronto, Canada. Association for Computational Linguistics.

\bibitem[{Lee et~al.(2022)Lee, Yang, Oh, and Seo}]{lee-etal-2022-generative}
Hyunji Lee, Sohee Yang, Hanseok Oh, and Minjoon Seo. 2022.
\newblock \href {https://doi.org/10.18653/v1/2022.emnlp-main.92} {Generative multi-hop retrieval}.
\newblock In \emph{Proceedings of the 2022 Conference on Empirical Methods in Natural Language Processing}, pages 1417--1436, Abu Dhabi, United Arab Emirates. Association for Computational Linguistics.

\bibitem[{Lee et~al.(2023{\natexlab{b}})Lee, Choi, and Lee}]{lee-etal-2023-glen}
Sunkyung Lee, Minjin Choi, and Jongwuk Lee. 2023{\natexlab{b}}.
\newblock \href {https://doi.org/10.18653/v1/2023.emnlp-main.477} {{GLEN}: Generative retrieval via lexical index learning}.
\newblock In \emph{Proceedings of the 2023 Conference on Empirical Methods in Natural Language Processing}, pages 7693--7704, Singapore. Association for Computational Linguistics.

\bibitem[{Li et~al.(2024{\natexlab{a}})Li, Dou, Zhou, and Liu}]{DBLP:conf/sigir/LiD0L24}
Xiaoxi Li, Zhicheng Dou, Yujia Zhou, and Fangchao Liu. 2024{\natexlab{a}}.
\newblock \href {https://doi.org/10.1145/3626772.3657778} {Corpuslm: Towards a unified language model on corpus for knowledge-intensive tasks}.
\newblock In \emph{Proceedings of the 47th International {ACM} {SIGIR} Conference on Research and Development in Information Retrieval, {SIGIR} 2024, Washington DC, USA, July 14-18, 2024}, pages 26--37. {ACM}.

\bibitem[{Li et~al.(2023{\natexlab{a}})Li, Yang, Wang, Wei, and Li}]{DBLP:journals/ipm/LiYWWL23}
Yongqi Li, Nan Yang, Liang Wang, Furu Wei, and Wenjie Li. 2023{\natexlab{a}}.
\newblock \href {https://doi.org/10.1016/J.IPM.2023.103475} {Generative retrieval for conversational question answering}.
\newblock \emph{Inf. Process. Manag.}, 60(5):103475.

\bibitem[{Li et~al.(2023{\natexlab{b}})Li, Yang, Wang, Wei, and Li}]{li-etal-2023-multiview}
Yongqi Li, Nan Yang, Liang Wang, Furu Wei, and Wenjie Li. 2023{\natexlab{b}}.
\newblock \href {https://doi.org/10.18653/v1/2023.acl-long.366} {Multiview identifiers enhanced generative retrieval}.
\newblock In \emph{Proceedings of the 61st Annual Meeting of the Association for Computational Linguistics (Volume 1: Long Papers)}, pages 6636--6648, Toronto, Canada. Association for Computational Linguistics.

\bibitem[{Li et~al.(2024{\natexlab{b}})Li, Yang, Wang, Wei, and Li}]{DBLP:conf/aaai/00010WWL24}
Yongqi Li, Nan Yang, Liang Wang, Furu Wei, and Wenjie Li. 2024{\natexlab{b}}.
\newblock \href {https://doi.org/10.1609/AAAI.V38I8.28717} {Learning to rank in generative retrieval}.
\newblock In \emph{Thirty-Eighth {AAAI} Conference on Artificial Intelligence, {AAAI} 2024, Thirty-Sixth Conference on Innovative Applications of Artificial Intelligence, {IAAI} 2024, Fourteenth Symposium on Educational Advances in Artificial Intelligence, {EAAI} 2014, February 20-27, 2024, Vancouver, Canada}, pages 8716--8723. {AAAI} Press.

\bibitem[{Li et~al.(2023{\natexlab{c}})Li, Zhang, Zhang, Long, Xie, and Zhang}]{li2023towards}
Zehan Li, Xin Zhang, Yanzhao Zhang, Dingkun Long, Pengjun Xie, and Meishan Zhang. 2023{\natexlab{c}}.
\newblock Towards general text embeddings with multi-stage contrastive learning.
\newblock \emph{arXiv preprint arXiv:2308.03281}.

\bibitem[{Ma et~al.(2021)Ma, Korotkov, Yang, Hall, and McDonald}]{ma2021zero}
Ji~Ma, Ivan Korotkov, Yinfei Yang, Keith Hall, and Ryan McDonald. 2021.
\newblock Zero-shot neural passage retrieval via domain-targeted synthetic question generation.
\newblock In \emph{Proceedings of the 16th Conference of the European Chapter of the Association for Computational Linguistics: Main Volume}, pages 1075--1088.

\bibitem[{Nogueira et~al.(2019)Nogueira, Lin, and Epistemic}]{nogueiradoc2query}
Rodrigo Nogueira, Jimmy Lin, and AI~Epistemic. 2019.
\newblock From doc2query to doctttttquery.

\bibitem[{Ouyang et~al.(2022)Ouyang, Wu, Jiang, Almeida, Wainwright, Mishkin, Zhang, Agarwal, Slama, Ray, Schulman, Hilton, Kelton, Miller, Simens, Askell, Welinder, Christiano, Leike, and Lowe}]{DBLP:conf/nips/Ouyang0JAWMZASR22}
Long Ouyang, Jeffrey Wu, Xu~Jiang, Diogo Almeida, Carroll~L. Wainwright, Pamela Mishkin, Chong Zhang, Sandhini Agarwal, Katarina Slama, Alex Ray, John Schulman, Jacob Hilton, Fraser Kelton, Luke Miller, Maddie Simens, Amanda Askell, Peter Welinder, Paul~F. Christiano, Jan Leike, and Ryan Lowe. 2022.
\newblock \href {http://papers.nips.cc/paper\_files/paper/2022/hash/b1efde53be364a73914f58805a001731-Abstract-Conference.html} {Training language models to follow instructions with human feedback}.
\newblock In \emph{Advances in Neural Information Processing Systems 35: Annual Conference on Neural Information Processing Systems 2022, NeurIPS 2022, New Orleans, LA, USA, November 28 - December 9, 2022}.

\bibitem[{Pang et~al.(2024)Pang, Yuan, Cho, He, Sukhbaatar, and Weston}]{DBLP:journals/corr/abs-2404-19733}
Richard~Yuanzhe Pang, Weizhe Yuan, Kyunghyun Cho, He~He, Sainbayar Sukhbaatar, and Jason Weston. 2024.
\newblock \href {https://doi.org/10.48550/ARXIV.2404.19733} {Iterative reasoning preference optimization}.
\newblock \emph{CoRR}, abs/2404.19733.

\bibitem[{Pradeep et~al.(2023)Pradeep, Hui, Gupta, Lelkes, Zhuang, Lin, Metzler, and Tran}]{pradeep-etal-2023-generative}
Ronak Pradeep, Kai Hui, Jai Gupta, Adam Lelkes, Honglei Zhuang, Jimmy Lin, Donald Metzler, and Vinh Tran. 2023.
\newblock \href {https://doi.org/10.18653/v1/2023.emnlp-main.83} {How does generative retrieval scale to millions of passages?}
\newblock In \emph{Proceedings of the 2023 Conference on Empirical Methods in Natural Language Processing}, pages 1305--1321, Singapore. Association for Computational Linguistics.

\bibitem[{Rafailov et~al.(2023)Rafailov, Sharma, Mitchell, Manning, Ermon, and Finn}]{DBLP:conf/nips/RafailovSMMEF23}
Rafael Rafailov, Archit Sharma, Eric Mitchell, Christopher~D. Manning, Stefano Ermon, and Chelsea Finn. 2023.
\newblock \href {http://papers.nips.cc/paper\_files/paper/2023/hash/a85b405ed65c6477a4fe8302b5e06ce7-Abstract-Conference.html} {Direct preference optimization: Your language model is secretly a reward model}.
\newblock In \emph{Advances in Neural Information Processing Systems 36: Annual Conference on Neural Information Processing Systems 2023, NeurIPS 2023, New Orleans, LA, USA, December 10 - 16, 2023}.

\bibitem[{Raffel et~al.(2020)Raffel, Shazeer, Roberts, Lee, Narang, Matena, Zhou, Li, and Liu}]{DBLP:journals/jmlr/RaffelSRLNMZLL20}
Colin Raffel, Noam Shazeer, Adam Roberts, Katherine Lee, Sharan Narang, Michael Matena, Yanqi Zhou, Wei Li, and Peter~J. Liu. 2020.
\newblock \href {https://jmlr.org/papers/v21/20-074.html} {Exploring the limits of transfer learning with a unified text-to-text transformer}.
\newblock \emph{J. Mach. Learn. Res.}, 21:140:1--140:67.

\bibitem[{Rasley et~al.(2020)Rasley, Rajbhandari, Ruwase, and He}]{DBLP:conf/kdd/RasleyRRH20}
Jeff Rasley, Samyam Rajbhandari, Olatunji Ruwase, and Yuxiong He. 2020.
\newblock \href {https://doi.org/10.1145/3394486.3406703} {Deepspeed: System optimizations enable training deep learning models with over 100 billion parameters}.
\newblock In \emph{{KDD} '20: The 26th {ACM} {SIGKDD} Conference on Knowledge Discovery and Data Mining, Virtual Event, CA, USA, August 23-27, 2020}, pages 3505--3506. {ACM}.

\bibitem[{Robertson and Walker(1994)}]{DBLP:conf/sigir/RobertsonW94}
Stephen~E. Robertson and Steve Walker. 1994.
\newblock \href {https://doi.org/10.1007/978-1-4471-2099-5\_24} {Some simple effective approximations to the 2-poisson model for probabilistic weighted retrieval}.
\newblock In \emph{Proceedings of the 17th Annual International {ACM-SIGIR} Conference on Research and Development in Information Retrieval. Dublin, Ireland, 3-6 July 1994 (Special Issue of the {SIGIR} Forum)}, pages 232--241. ACM/Springer.

\bibitem[{Sun et~al.(2023)Sun, Yan, Chen, Wang, Zhu, Ren, Chen, Yin, de~Rijke, and Ren}]{DBLP:conf/nips/0001YCWZRCYRR23}
Weiwei Sun, Lingyong Yan, Zheng Chen, Shuaiqiang Wang, Haichao Zhu, Pengjie Ren, Zhumin Chen, Dawei Yin, Maarten de~Rijke, and Zhaochun Ren. 2023.
\newblock \href {http://papers.nips.cc/paper\_files/paper/2023/hash/91228b942a4528cdae031c1b68b127e8-Abstract-Conference.html} {Learning to tokenize for generative retrieval}.
\newblock In \emph{Advances in Neural Information Processing Systems 36: Annual Conference on Neural Information Processing Systems 2023, NeurIPS 2023, New Orleans, LA, USA, December 10 - 16, 2023}.

\bibitem[{Tang et~al.(2024{\natexlab{a}})Tang, Chen, Li, Yu, Lu, Fu, Yu, Lin, Huang, He, Han, Sun, and Li}]{tang2024selfretrievalendtoendinformationretrieval}
Qiaoyu Tang, Jiawei Chen, Zhuoqun Li, Bowen Yu, Yaojie Lu, Cheng Fu, Haiyang Yu, Hongyu Lin, Fei Huang, Ben He, Xianpei Han, Le~Sun, and Yongbin Li. 2024{\natexlab{a}}.
\newblock \href {https://arxiv.org/abs/2403.00801} {Self-retrieval: End-to-end information retrieval with one large language model}.
\newblock \emph{Preprint}, arXiv:2403.00801.

\bibitem[{Tang and Yang(2024)}]{DBLP:journals/corr/abs-2401-15391}
Yixuan Tang and Yi~Yang. 2024.
\newblock \href {https://doi.org/10.48550/ARXIV.2401.15391} {Multihop-rag: Benchmarking retrieval-augmented generation for multi-hop queries}.
\newblock \emph{CoRR}, abs/2401.15391.

\bibitem[{Tang et~al.(2023)Tang, Zhang, Guo, Chen, Zhu, Wang, Yin, and Cheng}]{DBLP:conf/kdd/Tang0GCZWYC23}
Yubao Tang, Ruqing Zhang, Jiafeng Guo, Jiangui Chen, Zuowei Zhu, Shuaiqiang Wang, Dawei Yin, and Xueqi Cheng. 2023.
\newblock \href {https://doi.org/10.1145/3580305.3599903} {Semantic-enhanced differentiable search index inspired by learning strategies}.
\newblock In \emph{Proceedings of the 29th {ACM} {SIGKDD} Conference on Knowledge Discovery and Data Mining, {KDD} 2023, Long Beach, CA, USA, August 6-10, 2023}, pages 4904--4913. {ACM}.

\bibitem[{Tang et~al.(2024{\natexlab{b}})Tang, Zhang, Guo, de~Rijke, Chen, and Cheng}]{DBLP:journals/tois/TangZGRCC24}
Yubao Tang, Ruqing Zhang, Jiafeng Guo, Maarten de~Rijke, Wei Chen, and Xueqi Cheng. 2024{\natexlab{b}}.
\newblock \href {https://doi.org/10.1145/3653712} {Listwise generative retrieval models via a sequential learning process}.
\newblock \emph{{ACM} Trans. Inf. Syst.}, 42(5):133:1--133:31.

\bibitem[{Tay et~al.(2022)Tay, Tran, Dehghani, Ni, Bahri, Mehta, Qin, Hui, Zhao, Gupta, Schuster, Cohen, and Metzler}]{DBLP:conf/nips/Tay00NBM000GSCM22}
Yi~Tay, Vinh Tran, Mostafa Dehghani, Jianmo Ni, Dara Bahri, Harsh Mehta, Zhen Qin, Kai Hui, Zhe Zhao, Jai~Prakash Gupta, Tal Schuster, William~W. Cohen, and Donald Metzler. 2022.
\newblock \href {http://papers.nips.cc/paper\_files/paper/2022/hash/892840a6123b5ec99ebaab8be1530fba-Abstract-Conference.html} {Transformer memory as a differentiable search index}.
\newblock In \emph{Advances in Neural Information Processing Systems 35: Annual Conference on Neural Information Processing Systems 2022, NeurIPS 2022, New Orleans, LA, USA, November 28 - December 9, 2022}.

\bibitem[{Vaswani et~al.(2017)Vaswani, Shazeer, Parmar, Uszkoreit, Jones, Gomez, Kaiser, and Polosukhin}]{DBLP:conf/nips/VaswaniSPUJGKP17}
Ashish Vaswani, Noam Shazeer, Niki Parmar, Jakob Uszkoreit, Llion Jones, Aidan~N. Gomez, Lukasz Kaiser, and Illia Polosukhin. 2017.
\newblock \href {https://proceedings.neurips.cc/paper/2017/hash/3f5ee243547dee91fbd053c1c4a845aa-Abstract.html} {Attention is all you need}.
\newblock In \emph{Advances in Neural Information Processing Systems 30: Annual Conference on Neural Information Processing Systems 2017, December 4-9, 2017, Long Beach, CA, {USA}}, pages 5998--6008.

\bibitem[{von Werra et~al.(2020)von Werra, Belkada, Tunstall, Beeching, Thrush, Lambert, Huang, Rasul, and Gallouédec}]{vonwerra2022trl}
Leandro von Werra, Younes Belkada, Lewis Tunstall, Edward Beeching, Tristan Thrush, Nathan Lambert, Shengyi Huang, Kashif Rasul, and Quentin Gallouédec. 2020.
\newblock Trl: Transformer reinforcement learning.
\newblock \url{https://github.com/huggingface/trl}.

\bibitem[{Wang et~al.(2022)Wang, Thakur, Reimers, and Gurevych}]{wang2022gpl}
Kexin Wang, Nandan Thakur, Nils Reimers, and Iryna Gurevych. 2022.
\newblock Gpl: Generative pseudo labeling for unsupervised domain adaptation of dense retrieval.
\newblock In \emph{Proceedings of the 2022 Conference of the North American Chapter of the Association for Computational Linguistics: Human Language Technologies}, pages 2345--2360.

\bibitem[{Wang et~al.(2024)Wang, Yang, Huang, Yang, Majumder, and Wei}]{wang-etal-2024-improving-text}
Liang Wang, Nan Yang, Xiaolong Huang, Linjun Yang, Rangan Majumder, and Furu Wei. 2024.
\newblock \href {https://doi.org/10.18653/v1/2024.acl-long.642} {Improving text embeddings with large language models}.
\newblock In \emph{Proceedings of the 62nd Annual Meeting of the Association for Computational Linguistics (Volume 1: Long Papers)}, pages 11897--11916, Bangkok, Thailand. Association for Computational Linguistics.

\bibitem[{Wolf et~al.(2020)Wolf, Debut, Sanh, Chaumond, Delangue, Moi, Cistac, Rault, Louf, Funtowicz, Davison, Shleifer, von Platen, Ma, Jernite, Plu, Xu, Le~Scao, Gugger, Drame, Lhoest, and Rush}]{wolf-etal-2020-transformers}
Thomas Wolf, Lysandre Debut, Victor Sanh, Julien Chaumond, Clement Delangue, Anthony Moi, Pierric Cistac, Tim Rault, Remi Louf, Morgan Funtowicz, Joe Davison, Sam Shleifer, Patrick von Platen, Clara Ma, Yacine Jernite, Julien Plu, Canwen Xu, Teven Le~Scao, Sylvain Gugger, Mariama Drame, Quentin Lhoest, and Alexander Rush. 2020.
\newblock \href {https://doi.org/10.18653/v1/2020.emnlp-demos.6} {Transformers: State-of-the-art natural language processing}.
\newblock In \emph{Proceedings of the 2020 Conference on Empirical Methods in Natural Language Processing: System Demonstrations}, pages 38--45, Online. Association for Computational Linguistics.

\bibitem[{Xiao et~al.(2023)Xiao, Liu, Zhang, and Muennighoff}]{bge_embedding}
Shitao Xiao, Zheng Liu, Peitian Zhang, and Niklas Muennighoff. 2023.
\newblock \href {https://arxiv.org/abs/2309.07597} {C-pack: Packaged resources to advance general chinese embedding}.
\newblock \emph{Preprint}, arXiv:2309.07597.

\bibitem[{Yang et~al.(2023)Yang, Song, Zhang, Huang, Deng, Sun, and Zhang}]{yang-etal-2023-auto}
Tianchi Yang, Minghui Song, Zihan Zhang, Haizhen Huang, Weiwei Deng, Feng Sun, and Qi~Zhang. 2023.
\newblock \href {https://doi.org/10.18653/v1/2023.findings-emnlp.464} {Auto search indexer for end-to-end document retrieval}.
\newblock In \emph{Findings of the Association for Computational Linguistics: EMNLP 2023}, pages 6955--6970, Singapore. Association for Computational Linguistics.

\bibitem[{Yu et~al.(2023)Yu, Zhuang, Zhang, Meng, Ratner, Krishna, Shen, and Zhang}]{DBLP:conf/nips/YuZZMRKSZ23}
Yue Yu, Yuchen Zhuang, Jieyu Zhang, Yu~Meng, Alexander~J. Ratner, Ranjay Krishna, Jiaming Shen, and Chao Zhang. 2023.
\newblock \href {http://papers.nips.cc/paper\_files/paper/2023/hash/ae9500c4f5607caf2eff033c67daa9d7-Abstract-Datasets\_and\_Benchmarks.html} {Large language model as attributed training data generator: {A} tale of diversity and bias}.
\newblock In \emph{Advances in Neural Information Processing Systems 36: Annual Conference on Neural Information Processing Systems 2023, NeurIPS 2023, New Orleans, LA, USA, December 10 - 16, 2023}.

\bibitem[{Zeng et~al.(2024)Zeng, Luo, Jin, Sarwar, Wei, and Zamani}]{DBLP:conf/www/Zeng0JSWZ24}
Hansi Zeng, Chen Luo, Bowen Jin, Sheikh~Muhammad Sarwar, Tianxin Wei, and Hamed Zamani. 2024.
\newblock \href {https://doi.org/10.1145/3589334.3645477} {Scalable and effective generative information retrieval}.
\newblock In \emph{Proceedings of the {ACM} on Web Conference 2024, {WWW} 2024, Singapore, May 13-17, 2024}, pages 1441--1452. {ACM}.

\bibitem[{Zhang et~al.(2023)Zhang, Wang, Chen, Chang, Zhang, Miao, Hou, Ding, Miao, Wang, Pang, Zhan, Sun, Deng, Zhang, Yang, Xie, Yang, and Cui}]{DBLP:conf/nips/ZhangWCCZMHDMWP23}
Hailin Zhang, Yujing Wang, Qi~Chen, Ruiheng Chang, Ting Zhang, Ziming Miao, Yingyan Hou, Yang Ding, Xupeng Miao, Haonan Wang, Bochen Pang, Yuefeng Zhan, Hao Sun, Weiwei Deng, Qi~Zhang, Fan Yang, Xing Xie, Mao Yang, and Bin Cui. 2023.
\newblock \href {http://papers.nips.cc/paper\_files/paper/2023/hash/ac112e8ffc4e5b9ece32070440a8ca43-Abstract-Conference.html} {Model-enhanced vector index}.
\newblock In \emph{Advances in Neural Information Processing Systems 36: Annual Conference on Neural Information Processing Systems 2023, NeurIPS 2023, New Orleans, LA, USA, December 10 - 16, 2023}.

\bibitem[{Zhang et~al.(2024)Zhang, Liu, Zhou, Dou, Liu, and Cao}]{DBLP:conf/sigir/ZhangL0DLC24}
Peitian Zhang, Zheng Liu, Yujia Zhou, Zhicheng Dou, Fangchao Liu, and Zhao Cao. 2024.
\newblock \href {https://doi.org/10.1145/3626772.3657797} {Generative retrieval via term set generation}.
\newblock In \emph{Proceedings of the 47th International {ACM} {SIGIR} Conference on Research and Development in Information Retrieval, {SIGIR} 2024, Washington DC, USA, July 14-18, 2024}, pages 458--468. {ACM}.

\bibitem[{Zhao et~al.(2024)Zhao, Chen, Chen, Zhang, and Wu}]{DBLP:journals/corr/abs-2405-02714}
Xinran Zhao, Tong Chen, Sihao Chen, Hongming Zhang, and Tongshuang Wu. 2024.
\newblock \href {https://doi.org/10.48550/ARXIV.2405.02714} {Beyond relevance: Evaluate and improve retrievers on perspective awareness}.
\newblock \emph{CoRR}, abs/2405.02714.

\bibitem[{Zhou et~al.(2023)Zhou, Dou, and Wen}]{zhou-etal-2023-enhancing-generative}
Yujia Zhou, Zhicheng Dou, and Ji-Rong Wen. 2023.
\newblock \href {https://doi.org/10.18653/v1/2023.emnlp-main.768} {Enhancing generative retrieval with reinforcement learning from relevance feedback}.
\newblock In \emph{Proceedings of the 2023 Conference on Empirical Methods in Natural Language Processing}, pages 12481--12490, Singapore. Association for Computational Linguistics.

\bibitem[{Zhou et~al.(2022)Zhou, Yao, Dou, Wu, Zhang, and Wen}]{DBLP:journals/corr/abs-2208-09257}
Yujia Zhou, Jing Yao, Zhicheng Dou, Ledell Wu, Peitian Zhang, and Ji{-}Rong Wen. 2022.
\newblock \href {https://doi.org/10.48550/ARXIV.2208.09257} {Ultron: An ultimate retriever on corpus with a model-based indexer}.
\newblock \emph{CoRR}, abs/2208.09257.

\bibitem[{Zhuang et~al.(2022)Zhuang, Ren, Shou, Pei, Gong, Zuccon, and Jiang}]{DBLP:journals/corr/abs-2206-10128}
Shengyao Zhuang, Houxing Ren, Linjun Shou, Jian Pei, Ming Gong, Guido Zuccon, and Daxin Jiang. 2022.
\newblock \href {https://doi.org/10.48550/ARXIV.2206.10128} {Bridging the gap between indexing and retrieval for differentiable search index with query generation}.
\newblock \emph{CoRR}, abs/2206.10128.

\bibitem[{Ziems et~al.(2023)Ziems, Yu, Zhang, and Jiang}]{ziems-etal-2023-large}
Noah Ziems, Wenhao Yu, Zhihan Zhang, and Meng Jiang. 2023.
\newblock \href {https://doi.org/10.18653/v1/2023.findings-acl.167} {Large language models are built-in autoregressive search engines}.
\newblock In \emph{Findings of the Association for Computational Linguistics: ACL 2023}, pages 2666--2678, Toronto, Canada. Association for Computational Linguistics.

\end{thebibliography}
\appendix
\begin{table*}[htbp]
    \small
    \centering
    \begin{tabular}{lcccc}
    \toprule
    \multirow{2}{1cm}{\textbf{Dataset}} & \multirow{2}{1.1cm}{\textbf{Context}} & \multicolumn{3}{c}{\textbf{Queries}} \\
     & &\textbf{Chunk-Level} & \textbf{Sentence-Level} & \textbf{Constraints-Based}\\\midrule
     MultiHop-RAG & 7,724 & 72,090 & 472,193 & 51,212\\
     AllSides & 645 & 6,313 & 173,898 & 6,091 \\
     AGNews & 1,050 & 10,355 & 80,524 & 20,875 \\
     NQ & 98,748 & 1,459,031 & - & - \\\bottomrule %
    \end{tabular}
    \caption{Dataset Statistics}
    \label{tab:dataset_stats}
\end{table*}

\begin{table}[htbp]
    \small
    \centering
    \begin{tabular}{lp{4.8cm}}
    \toprule
      \textbf{Dataset} & \textbf{Attributes} \\\midrule
        MultiHop-RAG & author, publish time, source, category, title \\
        AllSides & political polarity \\
        AGNews & location, topic \\\bottomrule
    \end{tabular}
    \caption{Attributes used in each dataset for constraints-based query generation.}
    \label{tab:dataset_attributes}
\end{table}

\section{Details of Experiment Setup}
We use \texttt{Mistral-7B-Instruct-v0.3} as the base model for generative retrieval with the semantic identifier, while use \texttt{Mistral-7B-v0.3} as the base model for atomic identifier as it is closer to a classification setting.

For supervised fine-tuning, we train the models with 2 epochs, with a learning rate of 2e-5 and a warmup ratio of 0.1. The batch size is set as 256. We use sequence packing to put multiple examples in one forward pass~\citep{DBLP:journals/jmlr/RaffelSRLNMZLL20}. We use \texttt{bfloat16} for our training.

For preference learning, we mainly conduct experiments on MultiHop-RAG and NQ with semantic identifiers. We train the models with 1 epoch. The learning rate is set as 1e-7, batch size is set as 64, $\beta$ is set as 0.5, $\alpha$ is set as 1.0. 

The training infrastructure includes TRL~\citep{vonwerra2022trl}, Accelerate~\citep{accelerate}, Transformers~\citep{wolf-etal-2020-transformers}, DeepSpeed~\cite{DBLP:conf/kdd/RasleyRRH20} and FlashAttention-2~\citep{DBLP:conf/iclr/Dao24}. We use 8x Nvidia A100-SXM4-40GB for our experiments. Each training or inference procedure can be completed in 1 day.

Statistics of the numbers of the documents, different synthetic queries can be found in Table~\ref{tab:dataset_stats}. Attributes used for constraints-based synthetic queries can be found in Table~\ref{tab:dataset_attributes}. All the experiment results are obtained with single run.
\label{app:data_specific_setup}
\subsection{MultiHop-RAG}
On MultiHop-RAG, we split the documents into chunks with maximum length of 256 without overlap and conduct retrieval on individual chunks. For synthetic query generation, $m_c$, $m_s$ and $m_i$ are set as 10, and the temperature for LLM inference on synthetic data generation is set as 0.7. We interleave the Context2ID and Query2ID data as the full dataset for model supervised fine-tuning. The maximum sequence length is set as 700. For synthetic queries for preference learning, we ask the LLM to generate 10 queries. We perform the retrieval with beam size as 10 and retrieve the top-10 candidates for each query to construct the candidate pairs.
\subsection{AllSides}
On AllSides, we conduct document-level retrieval. For synthetic query generation, $m_c$, $m_s$ and $m_i$ are set as 10, and the temperature for LLM inference on synthetic data generation is set as 0.7. For Context2ID data, as there are some long documents in the corpus, we will split the long context into chunks with maximum length of 256 without overlap. The Context2ID data is constructed to use all chunks in the document to predict its corresponding document identifier. We interleave the Context2ID and Query2ID data as the full dataset for model supervised fine-tuning. The maximum sequence length is set as 700.
\subsection{AGNews}
On AllSides, we conduct document-level retrieval. For synthetic query generation, $m_c$, $m_s$ and $m_i$ are set as 10, and the temperature for LLM inference on synthetic data generation is set as 0.7. Queries constructed by \citet{DBLP:journals/corr/abs-2405-02714} uses two different perspectives. The first perspective is either the location of the desired news or the topic, while the second perspective is that the news is similar to another given news in the query. As we mentioned Section~\ref{sec:experiment_setup}, we replace the second perspective with the another field so that each query consists of both location and topic perspectives. The topic and location information used for instruction-based synthetic query generation is extracted with Mixtral 8x7b. We interleave the Context2ID and Query2ID data as the full dataset for model supervised fine-tuning. The maximum sequence length is set as 700.
\subsection{NQ}
On NQ, we conduct document-level retrieval. We use the document prefixes from~\cite{DBLP:conf/icml/KishoreWLAW23} to produce the semantic identifiers. For synthetic query generation, we perform truncation on pages when they are too long so that we always have at least 1024 token space for model output. We set $m_c$ as 15 and temperature as 0.7. We do not include sentence-level synthetic queries as the number of those queries are too large to be included in training within a reasonable time. Instead, we include sentence-level Context2ID as the approximation, and use the sentences from the document prefixes from~\cite{DBLP:conf/icml/KishoreWLAW23} to predict corresponding document identifiers. In NQ, we have high quality human annotated training queries, which we also include as part of the Query2ID data and therefore we do not include instruction-based synthetic queries. We concatenate the Context2ID and Query2ID data as the full dataset for model supervised fine-tuning, as interleaving will produce a much larger dataset that cannot be trained within a reasonable time. The maximum sequence length is set as 450. For synthetic queries for preference learning, we also perform truncation as for supervised fine-tuning, and ask the LLM to generate 10 queries. As the generated query number is quite large for inference, we use the first 2 generated queries for each documents for preference learning. We perform the retrieval with beam size as 10 and retrieve the top-10 candidates for each query to construct the candidate pairs.

\section{Results on Comparison with Off-The-Shelf Retrieval Models}
\label{sec:detailed_dense_comparison}
The detailed results for each dataset are shown in Table~\ref{tab:dense_retrieval_comparison}. We run the retrieval models on MultiHop-RAG, NQ and AGNews to collect the results, and adopt the results of AllSides from \citet{DBLP:journals/corr/abs-2405-02714}.

\begin{table*}[htbp]
    \centering
    \small
    \begin{subtable}[t]{\textwidth}
        \centering
        \begin{tabular}{lcccc}
        \toprule
        \textbf{Model} & \textbf{HIT@4} & \textbf{HIT@10} & \textbf{MAP@10} & \textbf{MRR@10}\\
         \midrule
         BM25 & 64.35 & 78.31 & \bf 26.30 & \bf 58.32 \\
         bge-large-en-v1.5 & 58.80 & 78.36 & 19.96 & 42.57 \\
         Contriever-msmarco & 55.25 & 75.08 & 19.28 & 40.69 \\
         E5-mistral-7b-instruct & 54.01 & 79.56 & 19.11 & 40.77 \\
         GTE-Qwen2-7B-instruct & 63.24 & 83.55 & 22.02 & 47.50 \\
         ours & \bf 71.88 & \bf 89.80& 26.23& 54.94\\
         \bottomrule
        \end{tabular}
        \caption{MultiHop-RAG}
    \end{subtable}
    \vspace{0.2cm}
    
    \begin{subtable}[t]{\textwidth}
        \centering
        \begin{tabular}{lcccc}
        \toprule
        \textbf{Model} & \textbf{HIT@1} & \textbf{HIT@5} & \textbf{HIT@10} & \textbf{MRR@10}\\
         \midrule
         BM25 & 32.82 & 53.70 & 60.92 & 42.45 \\
         bge-large-en-v1.5 & 55.59 & 76.58 & 81.75 & 64.45 \\
         Contriever-msmarco & 53.79 & 76.16 & 81.69 & 63.36 \\
         E5-mistral-7b-instruct & 59.07 & 80.08 & 85.28 & 68.11 \\
         GTE-Qwen2-7B-instruct & 60.45 & 80.87 & 85.72 & 69.30 \\
         ours & \bf 71.22& \bf 87.41& \bf 89.97& \bf 78.14\\
         \bottomrule
        \end{tabular}
        \caption{NQ}
    \end{subtable}
    \vspace{0.2cm}

    \begin{subtable}[t]{0.48\textwidth}
        \centering
        \begin{tabular}{lccc}
        \toprule
        \textbf{Model} & \textbf{HIT@1} & \textbf{HIT@5} & \textbf{HIT@10}\\
        \midrule
        BM25 & 5.86 & 26.85 & 36.42 \\
        bge-large-en-v1.5 & 6.94 & 27.32 & 34.11\\
        Contriever-msmarco & 6.64 & 25.77 & 38.43 \\
        E5-mistral-7b-instruct & 8.18 & 28.24 & 39.82\\
        GTE-Qwen2-7B-instruct & 9.11 & 34.11 & 49.07 \\
        ours & \bf 14.20& \bf 38.58& \bf 51.85\\
        \bottomrule
        \end{tabular}
        \caption{AllSides}
    \end{subtable}
    ~~\quad
    \begin{subtable}[t]{0.48\textwidth}
        \centering
        \begin{tabular}{lccc}
        \toprule
        \textbf{Model} & \textbf{HIT@1} & \textbf{HIT@5} & \textbf{HIT@10}\\
        \midrule
        BM25 & 38.70 & 67.47 & 77.63 \\
        bge-large-en-v1.5 & 54.14 & 80.57 & 86.53\\
        Contriever-msmarco & 52.69 & 80.40 & 85.79 \\
        E5-mistral-7b-instruct & 57.32 & \bf 85.90 & \bf 88.98 \\
        GTE-Qwen2-7B-instruct & 57.65 & 83.37 & 88.57\\
        ours & \bf 62.19& 83.78& 88.24\\
        \bottomrule
        \end{tabular}
        \caption{AGNews}
    \end{subtable}

    \caption{Comparisons to Off-The-Shelf Retrieval Models Across Datasets}
    \label{tab:dense_retrieval_comparison}
\end{table*}

\section{LLM Prompts}

\subsection{Prompts for Keywords Generation}
Figure~\ref{fig:keyword_generation} shows the prompt for generating a series of keywords as the semantic document identifier.

\begin{figure}[h]
\begin{tcolorbox}[title=\textbf{Keywords Generation Prompt}]
Summarize the following context with meaningful keywords representing different important information in the context. Your output should only consist a list of keywords in Markdown format, where each line starts with the dash "-" followed by the keywords.
\\
\\
\# Context:\\
\{context\}\\
\\
\# Keywords:
\end{tcolorbox}
\caption{Prompt for keywords-based document identifier generation.}
\label{fig:keyword_generation}
\end{figure}

\subsection{Prompts for Query Generation}
Figure~\ref{fig:query_generation_prompts} shows the prompts used to generate various types of synthetic queries, including chunk- and sentence-level queries, constructions-based queries, and question-answer pairs used at the preference learning stage.

\begin{figure*}[htbp]
\centering
\begin{subfigure}[t]{\textwidth}
\begin{tcolorbox}[title=\textbf{Query Generation Prompt}]
Your task is to generate a relevant and diverse set of \{num\_sequences\} questions that can be answered by the provided context. The questions are to be used by a retriever to retrieve the article from a large corpus. Your output should be a list of unordered in Markdown format, where each line starts with dash "-" followed by the question.
\\
\\
\# Context:\\
\{context\}\\
\\\
\# Output:
\end{tcolorbox}
\label{fig:query_generation}
\end{subfigure}

\begin{subfigure}[t]{\textwidth}
\begin{tcolorbox}[title=\textbf{Constraints-based Query Generation Prompt}]
Your task is to generate a diverse set of \{num\_sequences\} questions given a context with metadata. The generated questions should be answerable by the provided context. The questions are to be used by a retriever to retrieve the article from a large corpus. In addition, the question MUST be composed with at least one metadata filtering requirement.\\
\\
\textbf{\# MultiHop-RAG} \\
For example, if the source of the article is "New York Times", you can generate questions that specifically ask for certain information from "New York Times". You should generate questions with different metadata.\\
\textbf{\# AllSides and AGNews} \\
For example, if the source of the political polarity is "left", you can generate questions that specifically ask for certain information from "left-wing" source.\\
\\
DO NOT use "the context" or "the article" in any generated queries or answers.\\
DO NOT use pronoun "this" in any generated queries or answers.\\
DO NOT leak any information in this instruction.\\

Your output should be a list of unordered in Markdown format, where each line starts with dash "-" followed by the question. You do not need to provide the answer.\\
\\
\# Metadata\\
\{metadata\}\\
\\
\# Context\\
\{context\}\\
\\
\# Output:
\end{tcolorbox}
\label{fig:instruct_query_generation}
\end{subfigure}

\begin{subfigure}[t]{\textwidth}
\begin{tcolorbox}[title=\textbf{Query-Answer Pair Generation Prompt}]
Your task is to generate a relevant and diverse set of less than \{num\_sequences\} search engine query and answer pairs given a context.\\
The queries should be similar to what people use with search engine to find the given context from a large corpus. The answers are expeced to be a short phrase.\\
You should make the queries as difficult as possible, but they should be answerable by the given context.\\
\\
Do not use "the context" or "the article" in any generated queries or answers.\\
Do not use pronoun "this" in any generated queries or answers.\\
Do not leak any information in this instruction.\\
\\
Your output should be a list of unordered items in Markdown format, where each item starts with dash "-", followed by "Query:" and the generated query, and then "Answer:" with the corresponding answer.\\
\\
\# Context\\
\{context\}\\
\\
\# Output:
\end{tcolorbox}
\label{fig:qa_generation}
\end{subfigure}
\caption{Prompts for different types of synthetic query generation.}
\label{fig:query_generation_prompts}
\end{figure*}

\end{document}